\providecommand{\RALFinalMode}{1}
\providecommand{\RALPublicationMode}{0}
\providecommand{\RALCleanMode}{1}
\providecommand{\RALFinalRevisionMarked}{1}
\newif\ifRALmetadataready
\RALmetadatareadytrue

\newif\ifRALfinal
\ifnum\RALFinalMode=1\relax
  \RALfinaltrue
\else
  \RALfinalfalse
\fi

\newif\ifRALpublished
\ifnum\RALPublicationMode=1\relax
  \RALpublishedtrue
  \RALfinaltrue
\else
  \RALpublishedfalse
\fi

\newcommand{\RALPaperTitle}{Closed-Form 6-DoF Pose Estimation of Medical Tools via Gradiometer-Based Electromagnetic Localization System}
\newcommand{\RALShortTitle}{Closed-Form 6-DoF Medical Tool Pose Estimation via GELS}
\newcommand{\RALFirstAuthorSurname}{Wu}

\newcommand{\RALReceivedDate}{June 3, 2026}
\newcommand{\RALRevisedDate}{September 9, 2026}
\newcommand{\RALAcceptedDate}{October 27, 2026}
\newcommand{\RALFinalSubmissionMonthYear}{October, 2026}
\newcommand{\RALPublicationDate}{Month Day, Year}
\newcommand{\RALCurrentVersionDate}{Month Day, Year}
\newcommand{\RALDOI}{10.1109/LRA.XXXX.XXXXXXX}

\newcommand{\RALDecisionEditor}{Jessica Burgner-Kahrs}
\newcommand{\RALPublishedEditor}{J. Burgner-Kahrs}
\newcommand{\RALAssociateEditor}{TBD}
\newcommand{\RALCorrespondingAuthor}{Jinhui Zhang}
\newcommand{\RALFundingAcknowledgment}{This work was supported by the National Natural Science Foundation of China under Grant U22A2048, by the Beijing Natural Science Foundation under Grant 4264128, and by the BIT Research and Innovation Promoting Project under Grant 2025YCXZ008}
\newcommand{\RALAffiliationText}{Z.~Wu, J.~Luo, Y.~Pu, S.~Wei, Y.~Chen, and J.~Zhang are with the School of Automation, Beijing Institute of Technology, Beijing, 100081, China (e-mail: zhiweiwu.cn@outlook.com, luojiahao.edu@outlook.com, yubo.pu@bit.edu.cn, siyiwei@bit.edu.cn, yuankai.chen@bit.edu.cn, and zhangjinh@bit.edu.cn).}
\newcommand{\RALPublishedAffiliationText}{Zhiwei Wu, Jiahao Luo, Yubo Pu, Siyi Wei, Yuankai Chen, and Jinhui Zhang are with the School of Automation, Beijing Institute of Technology, Beijing, 100081, China (e-mail: zhiweiwu.cn@outlook.com, luojiahao.edu@outlook.com, yubo.pu@bit.edu.cn, siyiwei@bit.edu.cn, yuankai.chen@bit.edu.cn, and zhangjinh@bit.edu.cn).}
\newcommand{\RALDOIText}{Digital Object Identifier (DOI): see top of this page.}
\newcommand{\RALPublishedDOIText}{Digital Object Identifier \RALDOI}
\newcommand{\RALSupplementaryMaterialText}{This article has supplementary downloadable material available at https://doi.org/\RALDOI, provided by the authors.}
\newcommand{\RALRepositoryURL}{https://github.com/WayneWu0526/GELS_code}
\newcommand{\RALSupplementaryURL}{https://github.com/WayneWu0526/GELS_code/blob/main/Supplementary.pdf}

\newcommand{\RALFinalKeywords}{Medical Robots and Systems, Localization, electromagnetic localization, pose estimation.}

\ifRALfinal
  \documentclass[letterpaper,10pt,journal,twoside]{IEEEtran}
\else
  \documentclass[lettersize,journal]{IEEEtran}
\fi
\usepackage{graphicx}
\usepackage{picinpar}
\usepackage{url}
\usepackage{colortbl}
\usepackage{soul}
\usepackage{multirow}
\usepackage{pifont}
\usepackage{color}
\usepackage{alltt}
\usepackage[hidelinks]{hyperref}
\usepackage{enumerate}
\usepackage{siunitx}
\usepackage{epstopdf}
\usepackage{pbox}
\usepackage{enumitem}
\usepackage{needspace}

\usepackage{array}
\usepackage{stfloats}
\usepackage{verbatim}
\usepackage{graphicx}
\usepackage[nocompress]{cite}
\usepackage{dutchcal}
\usepackage{supertabular}%xtable
\usepackage[table,xcdraw]{xcolor}
\ifRALfinal
  \usepackage[commandnameprefix=ifneeded,final]{changes}
\else
  \ifnum\RALCleanMode=1\relax
    \usepackage[commandnameprefix=ifneeded,final]{changes}
  \else
    \usepackage[commandnameprefix=ifneeded,draft]{changes}
  \fi
\fi
\makeatletter
\AddToHook{cmd/added/before}{\def\Changes@AuthorColor{blue}}
\makeatother
\setcommentmarkup{\textsuperscript{#1}}
\newcommand{\revised}[1]{%
  {\ifnum\RALFinalRevisionMarked=1\relax\color{red}%
   \else\ifRALfinal\color{black}%
   \else\ifnum\RALCleanMode=1\relax\color{black}%
   \else\color{red}\fi\fi\fi#1}%
}
\newcommand{\revisedcalib}[1]{%
  {\ifnum\RALFinalRevisionMarked=1\relax\color{blue}%
   \else\ifRALfinal\color{black}%
   \else\ifnum\RALCleanMode=1\relax\color{black}%
   \else\color{blue}\fi\fi\fi#1}%
}
\usepackage{hyperref}
\usepackage{orcidlink}
\hypersetup{colorlinks=true,
	linkcolor=blue,
	urlcolor=blue,
	anchorcolor=blue,
	citecolor=blue}
\makeatletter
\renewcommand{\@cite}[2]{%
  {\color{blue}[#1\if@tempswa , #2\fi]}%
}
\makeatother
\usepackage{lipsum}
\usepackage{mwe}
\usepackage{makecell}
\usepackage{pifont}
\usepackage{nomencl}
\makenomenclature

\usepackage{amsmath,amsfonts,amssymb,amsthm,mathrsfs,mathtools}
\usepackage{bm}
\usepackage[linesnumbered,ruled,vlined]{algorithm2e}

\newcommand{\mr}[1]{\mathrm{#1}} 
\newcommand{\mc}[1]{\mathcal{#1}}
\newcommand{\mbc}[1]{\mathbcal{#1}}

\newcommand{\pr}[3]{\prescript{#1}{#2}{#3}}

\newcommand{\mrd}{\mathrm{d}}

\newcommand{\lt}{\left(}
\newcommand{\rt}{\right)}

\DeclareMathOperator{\tr}{tr}
\DeclareMathOperator{\vc}{vec}
\DeclareMathOperator{\mat}{mat}
\DeclareMathOperator{\col}{col}
\DeclareMathOperator{\rank}{rank}

\DeclareMathOperator{\proj}{Proj}
\DeclareMathOperator{\nullspace}{null}

\ifRALfinal
  \ifRALmetadataready\else
    \PackageWarningNoLine{ral-metadata}{Final metadata still contains pending date fields. Keep this PDF as a preview only}
  \fi
  \ifRALpublished
    \PackageWarningNoLine{ral-metadata}{IEEE production metadata is not assigned. This publication-layout entry is a preview only}
  \fi
\fi
\begin{document}
\ifRALfinal
  \title{\RALPaperTitle}
\else
  \title{\added[]{\RALPaperTitle}}
\fi

\author{
  \ifRALfinal\else\vskip 1em\fi
  Zhiwei Wu$^{\orcidlink{https://orcid.org/0000-0002-3957-3063}}$,~\IEEEmembership{Graduate Student Member,~IEEE,}
  Jiahao~Luo$^{\orcidlink{0009-0003-3197-1782}}$,
  Yubo~Pu$^{\orcidlink{0009-0002-6556-765X}}$,
  Siyi~Wei$^{\orcidlink{0009-0005-9967-4025}}$,
  Yuankai~Chen$^{\orcidlink{0009-0004-5618-845X}}$,
  and Jinhui~Zhang$^{\orcidlink{0000-0002-2405-894X}}$%
  \ifRALfinal
    \ifRALpublished
      \thanks{Received \RALReceivedDate; accepted \RALAcceptedDate. Date of publication \RALPublicationDate; date of current version \RALCurrentVersionDate. This article was recommended for publication by Associate Editor \RALAssociateEditor\ and Editor \RALPublishedEditor\ upon evaluation of the reviewers' comments. \RALFundingAcknowledgment. \textit{(Corresponding author: \RALCorrespondingAuthor.)}}%
      \thanks{\RALPublishedAffiliationText}%
      \thanks{\RALSupplementaryMaterialText}%
      \thanks{\RALPublishedDOIText}%
    \else
      \thanks{Manuscript received: \RALReceivedDate; Revised: \RALRevisedDate; Accepted: \RALAcceptedDate.}%
      \thanks{This paper was recommended for publication by Editor \RALDecisionEditor\ upon evaluation of the Associate Editor and Reviewers' comments. \RALFundingAcknowledgment. (Corresponding author: \RALCorrespondingAuthor.)}%
      \thanks{\RALAffiliationText}%
      \thanks{\RALDOIText}%
    \fi
  \else
    \thanks{Manuscript received \RALReceivedDate. \RALFundingAcknowledgment. (Corresponding author: \RALCorrespondingAuthor.)}%
    \thanks{\RALAffiliationText}%
    \thanks{Digital Object Identifier (DOI):}%
  \fi
}
\ifRALfinal
  \markboth{IEEE Robotics and Automation Letters. Preprint Version. Accepted \RALFinalSubmissionMonthYear}%
  {\RALFirstAuthorSurname\ \MakeLowercase{\textit{et al.}}: \RALShortTitle}
\fi


\maketitle

\begin{abstract}
Electromagnetic tracking is promising for remote navigation of endoluminal medical tools.
However, existing six-degree-of-freedom (6-DoF) pose recovery approaches often require pre-calibrated workspace field maps or iterative nonlinear optimization.
This letter presents a Gradiometer-Based Electromagnetic Localization System (GELS), a closed-form tracking framework that uses a compact magnetometer array as an embedded quasi-gradiometer to estimate local magnetic fields and gradient tensors.
These quantities are mapped by the Euler homogeneous relation to displacements between source and array, from which Procrustes registration over multiple sources recovers the array orientation and position using at least three noncollinear sources.
The algorithm requires known source positions and array geometry, but no pre-calibrated workspace field maps, initial pose guesses, or calibrated source magnetic moments.
The recovered array pose also enables a demonstration of sublevel dipole localization by serving as a mobile magnetic reference frame.
Benchtop experiments across sensor array configurations and excitation modes validate the proposed closed-form 6-DoF localization pipeline, with a median solver runtime of \SI{172.00}{\micro\second}.
A perturbation analysis identifies inter-sensor inconsistency and dipole model mismatch as dominant accuracy limits, informing future sensor array and source design to improve pose accuracy.
\end{abstract}


\section{Introduction}

\IEEEPARstart{E}{lectromagnetic}
navigation systems can remotely generate forces and torques on magnetically responsive devices for precision telesurgery \cite{nelsonRemoteMagneticNavigation2024}, vascular interventions \cite{hwangReviewMagneticActuation2020}, and targeted therapies \cite{landersClinicallyReadyMagnetic2025}. As these systems move toward closed-loop and partially automated operation, real-time six-degree-of-freedom (6-DoF) pose tracking becomes essential, particularly when continuous fluoroscopy is undesirable because of ionizing radiation \cite{duImageIntegratedMagneticActuation2023}.
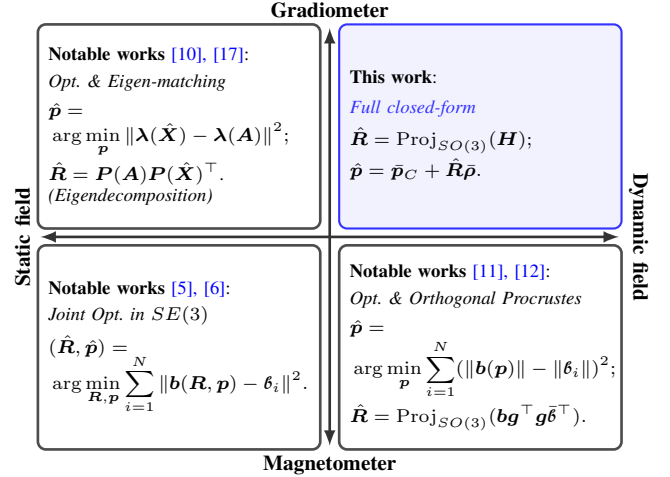
\begin{figure}[!t]
\centering
\newcommand{\figyes}{\textcolor{green!50!black}{\ding{51}}}
\newcommand{\figno}{\textcolor{red!80!black}{\ding{55}}}
\newcommand{\figunclear}{\textcolor{black!62}{$\triangle$}}
\newcommand{\figfio}[3]{%
  \textbf{F}:\,#1\hspace{0.5em}%
  \textbf{I}:\,#2\hspace{0.5em}%
  \textbf{O}:\,#3%
}
\begin{tikzpicture}[
    >=latex,
    line width=1pt,
    font=\small
]

% ================= 设置两种不同的 Box Style =================
% 基础背景框（用于其他人的工作）
\tikzset{
quadrant/.style={
    draw=black!70,          % 边框颜色稍微柔和一点
    fill=white,
    rounded corners=3pt,
    inner sep=4pt,
    align=left,
    text width=0.39\columnwidth, 
    minimum height=2.7cm,   % 强制统一高度！让四个框完全对称
    font=\scriptsize
}
}

% 高亮背景框（用于你的工作）
\tikzset{
highlight_quadrant/.style={
    quadrant,
    draw=blue!80,           % 用蓝色/主题色高亮边框
    thick,                  % 边框加粗
    fill=blue!5             % 极其淡的蓝色背景
},
fio_row/.style={
    anchor=south,
    font=\scriptsize,
    inner sep=0pt
}
}

% 整体坐标系尺寸系系数
\pgfmathsetlengthmacro{\L}{0.48*\columnwidth}

% 十字坐标轴
\draw[<->, draw=black!80] (-0.9*\L,0) -- (0.9*\L,0);
\draw[<->, draw=black!80] (0,-0.68*\L) -- (0,0.68*\L);

% 坐标标签 (稍微远离轴线一点防止与框干涉)
\node[rotate=90, anchor=south, font=\footnotesize\bfseries] at (-0.9*\L,0) {Static field};
\node[rotate=-90, anchor=south, font=\footnotesize\bfseries] at (0.9*\L,0) {Dynamic field};
\node[anchor=south, font=\footnotesize\bfseries] at (0,0.68*\L) {Gradiometer};
\node[anchor=north, font=\footnotesize\bfseries] at (0,-0.68*\L) {Magnetometer};

% ================= 坐标定义，保证绝对居中对称 =================
\def\PosX{0.225\columnwidth}
\def\PosY{0.165\columnwidth}

% -------- 第一象限 (Top-Right): Proposed --------
\node[highlight_quadrant, anchor=center] (tr)
at (\PosX, \PosY)
{
\textbf{This work (GELS):}\\[2pt]
\textit{\color{blue!80}Full closed-form}\\[3pt] 
% $\hat{\bm{\rho}}_i=-3\hat{\bm{X}}_i^{-1}\hat{\bar{\mbc{b}}}_i;$\\[2pt]
$\hat{\bm{R}}=\proj_{SO(3)}(\bm{H});$\\[2pt]
$\hat{\bm{p}}=\bar{\bm{p}}_C+\hat{\bm{R}}\bar{\bm{\rho}}.$
};

% -------- 第二象限 (Top-Left): Fischer --------
\node[quadrant, anchor=center] (tl)
at (-\PosX, \PosY)
{
\textbf{Notable work}
\cite{fischerGradiometerBasedMagneticLocalization2023}:\\[2pt]
\textit{Opt. \& Eigenvalue matching}\\[3pt]
$\begin{aligned}
&\hat{\bm{p}}=\\[-3pt]
&\arg\min_{\bm{p}}\|\bm\lambda(\hat{\bm{X}})-\bm\lambda(\bm{A})\|^2;
\end{aligned}$\\[2pt]
$\hat{\bm{R}}=\bm{P}(\bm{A})\bm{P}(\hat{\bm{X}})^\top$.
};

% -------- 第三象限 (Bottom-Left): Son --------
\node[quadrant, anchor=center] (bl)
at (-\PosX, -\PosY)
{
\textbf{Notable work} 
\cite{son5DLocalizationMethod2016}:\\[2pt]
\textit{Joint Opt. in $SE(3)$}\\[3pt]
$\begin{aligned}
    &(\hat{\bm{R}},\hat{\bm{p}})=\\[-5pt]
    &\arg\min_{\bm{R},\bm{p}}\sum_{i=1}^N\|\bm{b}(\bm{R},\bm{p})-\mbc{b}_i\|^2.
\end{aligned}$
};

% -------- 第四象限 (Bottom-Right): Arx --------
\node[quadrant, anchor=center] (br)
at (\PosX, -\PosY)
{
\textbf{Notable works} 
\cite{arxSimultaneousLocalizationActuation2023,vonarxOrthogonalPulseWidthModulationCombined2025}: \\[2pt]
\textit{Opt. \& Orthogonal Procrustes}\\[3pt]
$\begin{aligned}
&\hat{\bm{p}}=
\\[-3pt]
&\arg\min_{\bm{p}}\sum_{i=1}^M(\|\bm{b}(\bm{p})\|-\|\mbc{b}_i\|)^{2};
\end{aligned}$\\[2pt]
$\hat{\bm{R}}=\proj_{SO(3)}(\bm{b}\bm{g}^\top \bm{g}\bar{\mbc{b}}^\top).$\\[1mm]
\mbox{}
};

% Qualitative comparison, aligned at the bottom of each block.
\node[fio_row,yshift=1.2mm] at (tl.south) {\figfio{\figno}{\figno}{\figno}};
\node[fio_row,yshift=1.2mm] at (tr.south) {\figfio{\figyes}{\figyes}{\figyes}};
\node[fio_row,yshift=1.2mm] at (bl.south) {\figfio{\figno}{\figno}{\figno}};
\node[fio_row,yshift=1.2mm] at (br.south) {\figfio{\figno}{\figunclear}{\figno}};

\node[
  anchor=north,
  align=left,
  text width=0.94\columnwidth,
  font=\fontsize{6.4}{7.2}\selectfont,
  inner sep=0pt
] at (0,-0.84*\L) {%
  \textbf{F}: map free;\quad
  \textbf{I}: external initial pose guess free;\quad
  \textbf{O}: online iterative pose optimization free;\quad
  \figyes~= free; \figno~= not free; $\triangle$ = not explicit.
};

\end{tikzpicture}
\caption{Representative taxonomy of magnetic pose estimation approaches by field excitation and sensing modality.}
\label{fig:taxonomy}
\end{figure} Representative approaches include low field tracking with a 3-D magnetometer \cite{vergneLowFieldElectromagneticTracking2023}, 2-D Hall sensor arrays \cite{son5DLocalizationMethod2016}, external sensor arrays \cite{xuNovelSystemClosedLoop2021}, field model localization \cite{fischerUsingMagneticFields2022}, field map matching \cite{dinataliRealTimePoseDetection2013}, and magnetic gradient navigation \cite{sharmaWireless3DSurgical2021}. Embedded designs use magnetometers, pickup coils, or hybrid magnetic and inertial sensors \cite{arxSimultaneousLocalizationActuation2023,vonarxOrthogonalPulseWidthModulationCombined2025,popekSixDegreeofFreedomLocalizationUntethered2017}. Across these architectures, purely magnetic pointwise measurements couple position and orientation: position determines the spatial field sampled by the sensor, whereas orientation maps the same field into the sensor frame, so the two pose components cannot generally be recovered independently. Conventional methods therefore use nonlinear least squares optimization, pre-calibrated field maps, or auxiliary priors \cite{son5DLocalizationMethod2016,xuNovelSystemClosedLoop2021,fischerUsingMagneticFields2022,dinataliJacobianBasedIterativeMethod2016}. Inertial sensing can improve observability but introduces integration burden and drift in compact endoluminal devices \cite{daveigaSixDegreeFreedomLocalizationMultiple2023,daveigaMagneticLocalizationManipulation2025}.

Dynamic magnetic fields permit source contributions to be separated in time or frequency and have supported simultaneous actuation and localization \cite{arxSimultaneousLocalizationActuation2023,vonarxOrthogonalPulseWidthModulationCombined2025,taddeseEnhancedRealTimePose2018}. Source separation alone, however, does not resolve the pose inverse problem: the resulting pointwise field measurements still lack local spatial information, resulting in position estimation continues to rely on calibrated field maps or matching procedures with accurate prior models or initial guesses \cite{arxSimultaneousLocalizationActuation2023,vonarxOrthogonalPulseWidthModulationCombined2025}.

\begin{figure*}[ht]
    \centering
    \includegraphics[width=\textwidth]{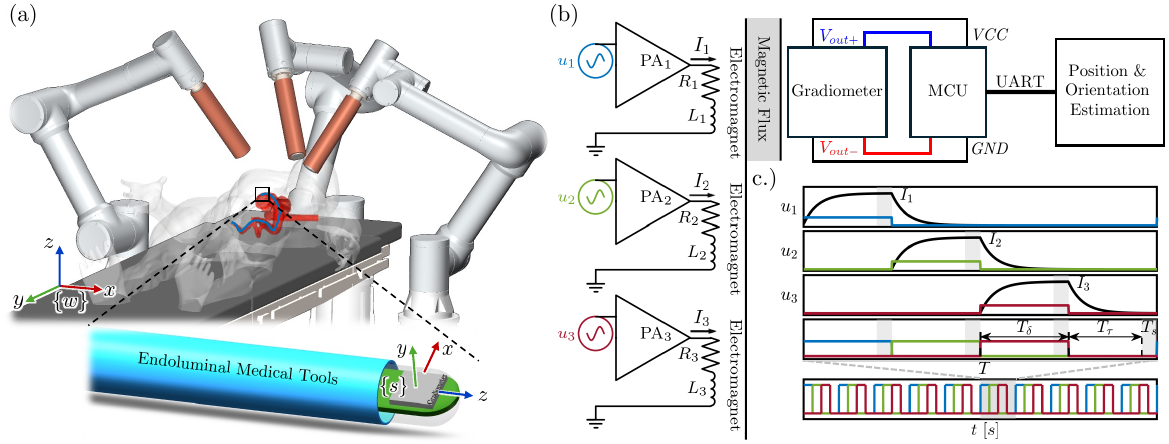}
    \caption{
    GELS for 6-DoF pose estimation in a conceptual endoluminal medical tool tracking scenario. 
    (a) System overview with multiple electromagnets and a medical tool carrying a compact magnetometer array. 
    (b) Signal flow from source excitation to MCU-based pose estimation. 
    (c) Time division multiplexing (TDM) sequence for isolating the contributions of individual sources.
    }
        \label{fig:system_overview}
\end{figure*}

Magnetic field gradients provide further local spatial structure and have supported gradient-based spatial encoding \cite{vergneLowFieldElectromagneticTracking2023,sharmaWireless3DSurgical2021}, scalar ranging \cite{beikiEstimatingSourceLocation2012,wiegertImprovedMagneticSTAR2008}, and closed-form dipole localization \cite{naraClosedFormFormulaMagnetic2006}. When the sensor pose is known, field and gradient measurements can be expressed in global coordinates, enabling analytical estimation without initial guesses or workspace field maps. For embedded gradiometers with unknown pose, however, the unknown sensor frame prevents direct use of the full directional gradient, so existing approaches still use pre-calibrated maps or optimization \cite{fischerGradiometerBasedMagneticLocalization2023,xiangPortableFlexibleIntermediary2026}. Gradient invariants can provide closed-form initial pose estimates, but their scalar range cues require at least four known sources and calibrated moments to recover absolute distance \cite{xiangPortableFlexibleIntermediary2026}, which is restrictive for electromagnets whose moments drift with coil heating \cite{sikorskiARMMSystemOptimized2019}.

These observations motivate the following question:
\textbf{can closed-form 6-DoF pose estimation of an embedded magnetometer array be achieved using purely magnetic measurements?}
We therefore propose a gradiometer-based electromagnetic localization system (GELS) that combines dynamic source separation with directional field and gradient information from an embedded quasi-gradiometer to achieve closed-form 6-DoF pose estimation. As summarized in Fig.~\ref{fig:taxonomy}, representative methods optimize pose in static fields or match dynamic fields to pre-calibrated maps, whereas GELS uses time division multiplexing (TDM) to isolate each source, estimates its local field and gradient, recovers the relative displacement to each source through the Euler homogeneous relation, and registers these displacements with known source positions through Orthogonal Procrustes. The localization priors comprise the source positions and array geometry, with no requirement for pre-calibrated workspace field maps, initial guesses, or calibrated source magnetic moments. We derive the uniqueness conditions, analyze errors arising from inter-sensor inconsistency and dipole model mismatch, validate the method on a benchtop system, and demonstrate sublevel dipole localization using the recovered array pose as a mobile magnetic reference frame.

\textbf{Notation}. Bold uppercase and lowercase letters denote matrices and vectors, respectively. The symbols $\|\cdot\|_2$ and $\|\cdot\|_F$ denote the Euclidean and Frobenius norms, while $\vc(\cdot)$ and $\mat(\cdot)$ denote vectorization and its inverse.
The operator $\col(\cdot)$ stacks its arguments vertically.
The symbols $(\cdot)^\dagger$, $\bm{I}$, and $\bm{1}$ denote the Moore–Penrose pseudoinverse, identity matrix, and column vector of ones, respectively.
The symbols $\{w\}$ and $\{s\}$ denote the world and sensor frames, respectively, where $\pr{w}{s}{\bm{R}}\in SO(3)$ maps vectors from $\{s\}$ to $\{w\}$, and $\pr{w}{}{\bm{p}}_s\in\mathbb{R}^3$ is the position of $\{s\}$ in $\{w\}$.

\section{Method}

\label{sec:method}

\subsection{Overview of GELS}

The proposed GELS is composed of $M$ electromagnets (Fig.~\ref{fig:system_overview}(a)), each with a known source position $\pr{w}{}{\bm{p}}_{C_k}$ and a dipole moment $\pr{w}{}{\bm{m}}_{C_k}$ that depends on excitation.
The system also includes a gradiometer rigidly mounted on the tracked medical tool, which sequentially estimates the magnetic quantities for each source $k=1,\dots,M$ from measurements. 
With the rigid transformation between the sensor and tool frames known, the recovered sensor pose directly determines the tool pose.
The quantities are composed of the local magnetic field $\pr{s}{}{\bar{\mbc{b}}}_k \in \mathbb{R}^3$ and the corresponding local gradient tensor $\pr{s}{}{\bm{X}}_k \in \mathbb{R}^{3\times 3}$. 
An MCU processes these measurements to provide the magnetic quantities for subsequent estimation (Fig.~\ref{fig:system_overview}(b)).

The input signals $\bm{u}_k(t)$ to each electromagnet are amplified by power amplifiers (PA$_k$) with gain $K$, producing coil currents $I_k(t)$, which generate magnetic fields at the sensor locations. 
Each dipole moment $\bm{m}_{C_k}$ is approximately proportional to the corresponding coil current $I_k(t)$ \cite{edelmannMagneticControlContinuum2017}.
The resulting magnetic field at a sensor position $\bm{p}$ can then be expressed as
\(
\pr{w}{}{\bm{b}}(\bm{p}) = \sum_k\bm{B}(\bm{p};\pr{w}{}{\bm{p}}_{C_k}) \, \pr{w}{}{\bm{m}}_{C_k},
\)
where $\bm{B}(\bm{p}) \in \mathbb{R}^{3 \times 3}$ represents the field contribution of each dipole moment at the sensor location \cite{abbott2020magnetic}.
This multi source field model underlies the localization pipeline summarized in Fig.~\ref{fig:overviewofpipline} and Algorithm~\ref{alg:supp_localization}, comprising time division excitation, local field and gradient estimation, and closed-form pose estimation.

\subsection{Source Separation via TDM Excitation}

The prototype employs a TDM excitation scheme to obtain measurements separated by source (Fig.~\ref{fig:system_overview}(c)).
Each source is sequentially excited by a rectangular pulse of width $T_{\delta}$.
To acquire measurements after the field settles, $T_{\delta} \geq T_{\tau}+T_s$, where $T_s$ is the gradiometer sampling period and $T_{\tau}$ is the current settling time.

The required $T_{\tau}$ depends on the amplifier mode.
In constant voltage (CV) mode, the coil current follows the first-order RL response, yielding $L \frac{\mrd I_k}{\mrd t}+RI_k=Ku_k(t)$, where $L$, $R$, and $\tau=L/R$ denote the coil inductance, resistance, and characteristic time constant, respectively.
Setting $T_{\tau}=5\tau$ allows the transient to decay but lowers the localization frequency.
In constant current (CC) mode, closed-loop regulation shortens the transient, allowing $T_{\tau}$ to be set well below $5\tau$, so $T_s$ primarily determines the localization frequency.
With one inactive slot for the background field $(\pr{s}{}{\bm{X}}_0,\pr{s}{}{\bar{\mbc{b}}_0})$, the multiplexing cycle satisfies $T\geq(M+1)T_{\delta}$, whereas without this slot, $T\geq MT_{\delta}$.

\begin{figure}[t]
    \centering
    \includegraphics[width=\columnwidth]{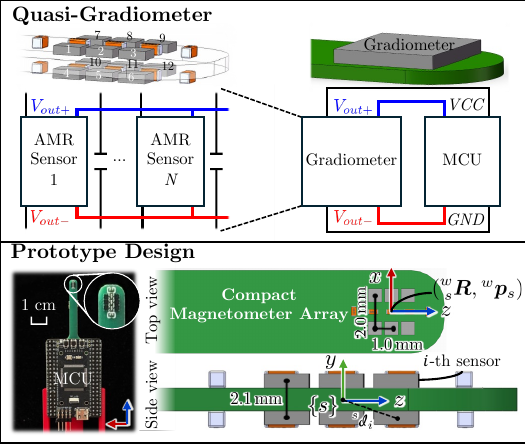}
    \caption{Quasi-gradiometer implementation using a compact magnetometer array. The top panel shows the gradiometer concept with multiple sensors and MCU connection, while the bottom panel shows the prototype geometry, sensor spacing, and center pose $(\prescript{w}{s}{\bm{R}}, \prescript{w}{}{\bm{p}}_s)$.}
    \label{fig:3}
\end{figure}

\subsection{Measurements via Quasi-Gradiometer}
\label{subsec:quasigradiometer}

We extract the local magnetic quantities using a compact magnetometer array configured as a quasi-gradiometer (Fig.~\ref{fig:3}).
The array contains $N$ three-axis anisotropic magnetoresistive (AMR) magnetometers rigidly mounted at known offsets $\prescript{s}{}{\mbc{d}_i} \in \mathbb{R}^3$ in the sensor frame $\{s\}$, producing measurements $\prescript{s}{}{\mbc{b}_i} \in \mathbb{R}^3$.
The predicted measurement at the $i$-th magnetometer is given by
\(
\prescript{s}{}{\mbc{b}_i} =
\prescript{w}{s}{\bm{R}}^\top
\sum_k
\prescript{w}{}{\bm{b}}_k
\big(
\prescript{w}{}{\bm{p}}_s + \prescript{w}{s}{\bm{R}}\prescript{s}{}{\mbc{d}_i};
\prescript{w}{}{\bm{m}_{C_k}}, \prescript{w}{}{\bm{p}_{C_k}}
\big).
\)
Hereafter, frame superscripts are omitted below when unambiguous, while calligraphic notation identifies local quantities.

\subsubsection{Local Magnetic Field Estimator}

The sensor offsets $\mbc{d}_i$ are small relative to the field variation scale, enabling the field at each sensor to be approximated to first order about the array center $\bm{p}$ as $\bm{b}(\bm{p}+\bm{R}\mbc{d}_i)\approx\bm{b}(\bm{p})+\bm{A}(\bm{p})\bm{R}\mbc{d}_i$, where $\bm{A}(\bm{p})\triangleq\nabla\bm{b}(\bm{p})$. In a source-free sensing region, $\nabla\cdot\bm{b}=0$ and $\nabla\times\bm{b}=\bm{0}$, so $\bm{A}(\bm{p})$ is symmetric and traceless.

Substituting the above expansion into the measurement model yields
\begin{equation}
\label{eq:linearized_measurement}
\mbc{b}_i
\approx
\bm{R}^\top\bm{b}(\bm{p})
+
\bm{R}^\top\bm{A}(\bm{p})\bm{R}\mbc{d}_i.
\end{equation}
Define the local magnetic quantities as $\bar{\mbc{b}}(\bm{p})\triangleq\bm{R}^\top\bm{b}(\bm{p})$ and $\bm{X}(\bm{p})\triangleq\bm{R}^\top\bm{A}(\bm{p})\bm{R}$, and stack the sensor measurements and offsets by columns as $\mbc{B}\triangleq[\mbc{b}_1,\dots,\mbc{b}_N]\in\mathbb{R}^{3\times N}$ and $\mbc{D}\triangleq[\mbc{d}_1,\dots,\mbc{d}_N]\in\mathbb{R}^{3\times N}$, respectively. The linearized model becomes
\begin{equation}
\label{eq:matrix_linearized_measurement}
\mbc{B}
\approx
\bar{\mbc{b}}(\bm{p})\bm{1}^\top
+
\bm{X}(\bm{p})\mbc{D}.
\end{equation}

We seek a weight vector $\bm{w}\in\mathbb{R}^N$ satisfying $\mbc{D}\bm{w}=\bm{0}$ and $\bm{1}^\top\bm{w}=1$ to eliminate the first-order gradient contribution. 
Right multiplying \eqref{eq:matrix_linearized_measurement} by $\bm{w}$ gives $\mbc{B}\bm{w}\approx\bar{\mbc{b}}(\bm{p})$, yielding
\(
\hat{\bar{\mbc{b}}}(\bm{p})=\mbc{B}\bm{w}.
\)
Thus, the estimate is a weighted average invariant to the first-order gradient term.
Let $r=\rank(\mbc{D})$, and let $\bm{Q}\in\mathbb{R}^{N\times(N-r)}$ contain an orthonormal basis of $\nullspace(\mbc{D})$, so $\mbc{D}\bm{Q}=\bm{0}$. Defining $\bm{g}\triangleq\bm{Q}^\top\bm{1}\in\mathbb{R}^{N-r}$, the choice
\(
\bm{w}
=
\bm{Q}\bm{g}/\|\bm{g}\|_2^2
\)
satisfies $\mbc{D}\bm{w}=\bm{0}$ and $\bm{1}^\top\bm{w}=1$, provided that $\|\bm{g}\|_2\neq 0$. 
We obtain the estimate
\begin{equation}
\label{eq:local_field_estimator}
\hat{\bar{\mbc{b}}}(\bm{p})
=\mbc{B}\bm{w}=
\mbc{B}\frac{\bm{Q}\bm{g}}{\|\bm{g}\|_2^2}.
\end{equation}
For a centered array $(\mbc{D}\bm{1}=\bm{0})$, \eqref{eq:local_field_estimator} reduces to $\hat{\bar{\mbc{b}}}(\bm{p})=\frac{1}{N}\mbc{B}\bm{1}$, so the formulation generalizes centroid averaging to arbitrary sensor layouts.

\begin{figure*}[ht]
    \centering
    \includegraphics[width=\linewidth]{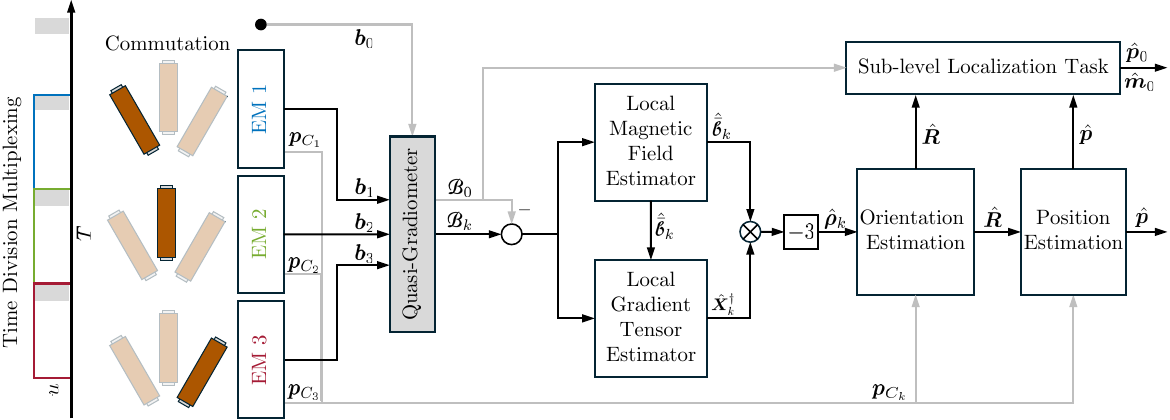}
    \caption{
        Overview of the localization pipeline. The TDM scheme sequentially excites each electromagnet (EM$_k$), and the quasi-gradiometer estimates $\hat{\bar{\mbc{b}}}_k$ and $\hat{\bm{X}}_k$ from the resulting measurements. These estimates yield relative displacements $\hat{\bm{\rho}}_k$ and the closed-form array pose $(\hat{\bm{R}},\hat{\bm{p}})$, which supports sublevel localization of an external dipole $(\hat{\bm{p}}_0,\hat{\bm{m}}_0)$.
    }
    \label{fig:overviewofpipline}
\end{figure*}

\subsubsection{Local Gradient Tensor Estimator}

Using the same weight vector $\bm{w}$, define the generalized centering matrix $\bm{P}_{\bm{w}}\triangleq\bm{I}_N-\bm{w}\bm{1}^\top$. Since $\mbc{D}\bm{w}=\bm{0}$ and $\bm{1}^\top\bm{w}=1$, $\bm{1}^\top\bm{P}_{\bm{w}}=\bm{0}^\top$ and $\mbc{D}\bm{P}_{\bm{w}}=\mbc{D}$. Right multiplying \eqref{eq:matrix_linearized_measurement} by $\bm{P}_{\bm{w}}$ eliminates the common local field term while preserving the first-order gradient contribution, yielding
\(
\mbc{B}\bm{P}_{\bm{w}} \approx \bm{X}(\bm{p})\mbc{D}.
\)
Since the local gradient tensor $\bm{X}$ is orthogonally similar to $\bm{A}(\bm{p})$, it remains symmetric and traceless. Accordingly, $\bm{X}$ can be estimated by solving
\begin{equation}
\label{eq:LSforX_letter}
\min_{\bm{X}}
\|\bm{X}\mbc{D}-\mbc{B}\bm{P}_{\bm{w}}\|_F^2
\quad\mr{s.t.}\ 
\bm{X}=\bm{X}^\top,\ \tr(\bm{X})=0.
\end{equation}

To enforce these structural constraints explicitly, we parameterize $\bm{X}$ as
\(
\bm{X}
=
\begin{bmatrix}
x_1 & x_2 & x_3\\
x_2 & x_4 & x_5\\
x_3 & x_5 & -(x_1+x_4)
\end{bmatrix},
\)
where $\bm{x}\triangleq[x_1,x_2,x_3,x_4,x_5]^\top\in\mathbb{R}^5$.
Let $\bm{S}\in\mathbb{R}^{9\times 5}$ be the constant selection matrix satisfying $\vc(\bm{X})=\bm{S}\bm{x}$. Using the identity $\vc(\bm{X}\mbc{D})=(\mbc{D}^\top\otimes\bm{I}_3)\vc(\bm{X})$, problem \eqref{eq:LSforX_letter} is converted into the linear least-squares problem $\min_{\bm{x}}\|\bm{C}\bm{x}-\bm{h}\|_2^2$, where $\bm{C}\triangleq(\mbc{D}^\top\otimes\bm{I}_3)\bm{S}$ and $\bm{h}\triangleq\vc\!\big(\mbc{B}\bm{P}_{\bm{w}}\big)$. If $\rank(\bm{C})=5$, the minimizer is unique and is given by $\hat{\bm{x}}=\bm{C}^\dagger\bm{h}$. Since $\bm C\bm x=\vc(\bm X\mbc D)$, $\rank(\bm C)=5$ if and only if $\rank(\mbc D)\geq2$. With two independent sensor offset directions, $\bm X\mbc D=\bm0$ forces $\bm X=\bm0$ under the symmetric and traceless constraints. Furthermore, $\mbc D\bm w=\bm0$ and $\bm1^\top\bm w=1$ imply $\rank(\mbc D)\leq N-1$, so at least three $(N\geq3)$ noncollinear magnetometers are necessary for unique estimation.
Beyond rank, the conditioning of $\bm C$ and sensor redundancy determine noise and offset amplification.
The local gradient tensor estimate is then reconstructed as
\begin{equation}
\label{eq:X_reconstruction}
\hat{\bm{X}}
=
\mat(\bm{S}\hat{\bm{x}})
=
\mat\!\left(\bm{S}\bm{C}^\dagger\vc\!\big(\mbc{B}\bm{P}_{\bm{w}}\big)\right).
\end{equation}
Together, the local magnetic quantities estimated in \eqref{eq:local_field_estimator} and \eqref{eq:X_reconstruction} provide the inputs to the closed-form pose estimator.

\begin{algorithm}[t]
\caption{Pose Estimation from Measurements Separated by Source}
\label{alg:supp_localization}
\KwIn{$N$, $M$, $\mbc{D}$, $\{\bm{p}_{C_k}\}_{k=1}^M$, $\{\mbc{B}_k\}_{k=1}^M$}
\KwOut{$(\hat{\bm{R}},\hat{\bm{p}})$}

Compute $\bm{Q}$ such that $\mbc{D}\bm{Q}=\bm{0}$, and set $\bm{g}\leftarrow\bm{Q}^\top\bm{1}$\;
Set $\bm{w}\leftarrow \bm{Q}\bm{g}/\|\bm{g}\|_2^2$ and $\bm{P}_{\bm{w}}\leftarrow \bm{I}_N-\bm{w}\bm{1}^\top$\;
Set $\bm{S}$ by $\vc(\bm{X})=\bm{S}\bm{x}$ and $\bm{C}\leftarrow(\mbc{D}^\top\otimes\bm{I}_3)\bm{S}$\;

\For{$k= 1$ \KwTo $M$}{
    $\hat{\bar{\mbc{b}}}_k \leftarrow \mbc{B}_k\bm{w}$\;
    $\hat{\bm{X}}_k \leftarrow \mat\!\left(\bm{S}\bm{C}^\dagger \vc(\mbc{B}_k\bm{P}_{\bm{w}})\right)$\;
    $\hat{\bm{\rho}}_k \leftarrow -3\hat{\bm{X}}_k^\dagger\hat{\bar{\mbc{b}}}_k$\;
}
Set
$\bar{\bm{p}}_C \leftarrow \frac{1}{M}\sum_{k=1}^{M}\bm{p}_{C_k}$ and
$\bar{\bm{\rho}} \leftarrow \frac{1}{M}\sum_{k=1}^{M}\hat{\bm{\rho}}_k$\;
\For{$k=1$ \KwTo $M$}{
    $\tilde{\bm{p}}_{C_k} \leftarrow \bm{p}_{C_k}-\bar{\bm{p}}_C$\;
    $\tilde{\bm{\rho}}_k \leftarrow \hat{\bm{\rho}}_k-\bar{\bm{\rho}}$\;
}
$\bm{H} \leftarrow \sum_{k=1}^{M}\tilde{\bm{p}}_{C_k}(-\tilde{\bm{\rho}}_k)^\top$\;
Compute $\bm{H}=\bm{U}\bm{\Sigma}\bm{V}^\top$ via SVD\;
$\hat{\bm{R}}\leftarrow\bm{U}\,\mathrm{diag}\!\left(1,1,\det(\bm{U}\bm{V}^\top)\right)\bm{V}^\top$\;
\tcp{$\hat{\bm{R}}=\proj_{SO(3)}(\bm{H})$}
$\hat{\bm{p}}\leftarrow \bar{\bm{p}}_C+\hat{\bm{R}}\bar{\bm{\rho}}$\;
\Return{$(\hat{\bm{R}},\hat{\bm{p}})$}\;
\end{algorithm}

\subsection{Closed-Form Pose Estimation}
For each first-order magnetic dipole source \cite{abbott2020magnetic}, the field and its gradient satisfy the Euler homogeneous relation \cite{naraClosedFormFormulaMagnetic2006}
\begin{equation}
\label{eq:euler_global_cf}
\bm{A}_k(\bm{p})(\bm{p}-\bm{p}_{C_k})=-3\bm{b}_k(\bm{p}).
\end{equation}
The effect of deviations from this ideal dipole assumption is analyzed later in Sec.~\ref{sec:model_mismatch}.
Transforming \eqref{eq:euler_global_cf} into the sensor frame and defining the relative displacement $\bm{\rho}_k \triangleq \bm{R}^\top (\bm{p}-\bm{p}_{C_k})$ yields
\(
\bm{X}_k \bm{\rho}_k = -3 \bar{\mbc{b}}_k.
\)
Then the relative displacement vector is estimated by
\begin{equation}
\hat{\bm{\rho}}_k=-3\hat{\bm{X}}_k^\dagger \hat{\bar{\mbc{b}}}_k,
\end{equation}
which reduces to $\hat{\bm{\rho}}_k=-3\hat{\bm{X}}_k^{-1} \hat{\bar{\mbc{b}}}_k$ when $\hat{\bm{X}}_k$ is nonsingular.
One can also implement a formulation without matrix inversion that improves numerical stability \cite{yinClosedformFormulaMagnetic2020}.
Notably, for each isolated source contribution, $\hat{\bar{\mbc{b}}}_k$ and $\hat{\bm{X}}_k$ share the same sensor gain or source strength scaling, so the ratio structure makes $\hat{\bm{\rho}}_k$ independent of both uniform gain and the external source magnetic moment.

The array pose is recovered by treating $\hat{\bm{\rho}}_k=\bm{R}^\top(\bm{p}-\bm{p}_{C_k})$ as a standard rigid registration problem on $SE(3)$ \cite{horn1987closed}, equivalently written as $\bm{p}_{C_k}=\bm{p}-\bm{R}\bm{\rho}_k$. Let $\bar{\bm{p}}_C \triangleq \frac{1}{M}\sum_{k=1}^M \bm{p}_{C_k}$ and $\bar{\bm{\rho}} \triangleq \frac{1}{M}\sum_{k=1}^M \hat{\bm{\rho}}_k$, and define the centered variables $\tilde{\bm{p}}_{C_k}\triangleq \bm{p}_{C_k}-\bar{\bm{p}}_C$ and $\tilde{\bm{\rho}}_k\triangleq \hat{\bm{\rho}}_k-\bar{\bm{\rho}}$. The rotation is obtained by the Kabsch solution of the induced Orthogonal Procrustes problem \cite{schonemann1966generalized}. Specifically, with
\(
\label{eq:cross_covariance_cf_short}
\bm{H}\triangleq \sum_{k=1}^{M}\tilde{\bm{p}}_{C_k}(-\tilde{\bm{\rho}}_k)^\top,
\)
and
\(
\bm{H}=\bm{U}\bm{\Sigma}\bm{V}^\top,
\)
the optimal estimate of orientation is
\begin{equation}
\label{eq:kabsch_standard_cf_short}
\hat{\bm{R}}
=
\bm{U}\,\mathrm{diag}\bigl(1,1,\det(\bm{U}\bm{V}^\top)\bigr)\bm{V}^\top.
\end{equation}
The uniqueness condition for this Procrustes problem is that the centered source geometry provides at least two independent directions. 
Thus, at least three noncollinear sources ($M\ge 3$) are required.
Finally, the array position is computed from the estimated orientation and relative displacements:
\begin{equation}
\label{eq:position_from_centroids_cf_short}
\hat{\bm{p}} = \bar{\bm{p}}_C + \hat{\bm{R}}\bar{\bm{\rho}}
= \frac{1}{M}\sum_{k=1}^{M}\left(\bm{p}_{C_k}+\hat{\bm{R}}\hat{\bm{\rho}}_k\right).
\end{equation}

\subsection{Sublevel Dipole Localization Task}
\label{subsec:sublevel}

Here, sublevel dipole localization denotes a task in which GELS first recovers the array pose and then uses it as the reference frame for estimating a magnetic target.
Once the array $(\hat{\bm{R}},\hat{\bm{p}})$ is known, the quasi-gradiometer can further estimate the position and magnetic moment of an unknown external dipole producing a measurable signal.
Let the external dipole be located at $\bm{p}_0$ with magnetic moment $\bm{m}_0$ and generate a field at the sensor positions modeled by
\(
\bm{b}_0 = \bm{B}(\bm{p};\bm{p}_0) \, \bm{m}_0
\).
The quasi-gradiometer provides the local measurements $(\hat{\bm{X}}_0, \hat{\bar{\mbc{b}}}_0)$ in the sensor frame. 
Applying the Euler homogeneous relation \eqref{eq:euler_global_cf} to the unknown dipole yields
\begin{equation}
\label{eq:sublevel_position}
\hat{\bm{p}}_0 = \hat{\bm{p}} -  \hat{\bm{R}}\hat{\bm{\rho}}_0,
\end{equation}
in which $\hat{\bm{\rho}}_0=-3\hat{\bm{X}}_0^{\dagger}\hat{\bar{\mbc{b}}}_0$.
With the position $\hat{\bm{p}}_0$ known, the dipole moment is recovered using the dipole field relation:
\begin{equation}
\hat{\bm{m}}_0=\bm{B}(\hat{\bm{p}};\hat{\bm{p}}_0)^{\dagger}\hat{\bm{R}}\hat{\bar{\mbc{b}}}_0.
\end{equation}

\section{Experiment}
\label{sec:experiment}

\begin{figure}[t]
    \centering
    \includegraphics[width=\columnwidth]{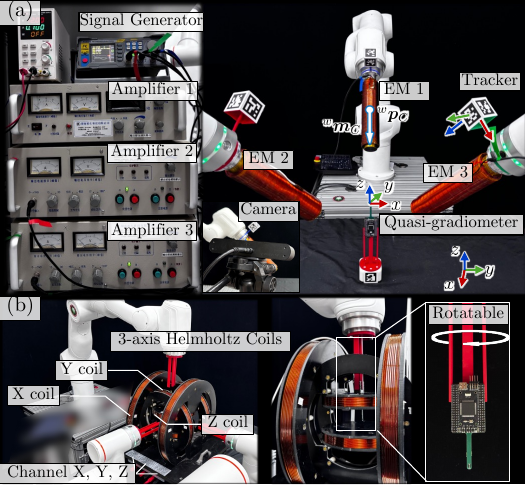}
    \caption{
    (a) Experimental setup for 6-DoF pose estimation. (b) Calibration setup for reducing inter-sensor measurement inconsistency.
    }
    \label{fig:expsetup}
\end{figure}

\subsection{Experimental Setup}

The 6-DoF experimental setup is shown in Fig.~\ref{fig:expsetup}(a).
GELS uses three electromagnets ($M=3$). Each electromagnet has an inductance of \SI{24.1}{\milli\henry} and a resistance of \SI{2.1}{\ohm}, is driven by a PSHEA-200C power amplifier, and is controlled by an STM32H743VIT6 MCU. 
The embedded quasi-gradiometer uses $N=12$ three-axis AMR sensors (QMC6309, QST Corporation), each with a $0.8 \times 0.8 \times 0.5\si{\milli\meter\cubed}$ WLCSP package, a $\pm3.2~\si{\milli\tesla}$ range, and a \SI{0.1}{\micro\tesla}--\SI{0.3}{\micro\tesla} resolution, with the geometry shown in Fig.~\ref{fig:3}. 
Because the MCU sequentially acquires the $N$ sensors through I2C buses at $f_s=\SI{1000}{\hertz}$, the gradiometer sampling period is $T_s=N/f_s=N~\si{\milli\second}$.

For the CV tests, the hardware enforces a dead time of $5\tau \approx\SI{57.38}{\milli\second}$ (with measured $\tau = \SI{11.48}{\milli\second}$) before sampling. 
The aggregated data are transmitted to a host PC equipped with an Intel Core i5-12490F CPU (3.0 GHz base frequency) and 32 GB of RAM, where the closed-form pose solver is executed in Python.
The solver achieves a median execution time of \SI{172.00}{\micro\second}, a minimum of \SI{151.10}{\micro\second}, and an average of \SI{186.66}{\micro\second}.

In the hardware configuration, the three electromagnets ($\mbox{core length}=250\ \mr{mm}$, $\mbox{radius}=20\ \mr{mm}$) are positioned by three robotic arms.
The equivalent dipole position $\bm{p}_{C_k}$ of each electromagnet is obtained using a ZED2i stereo camera tracking an AprilTag marker (precision $\approx \SI{0.5}{\milli\meter}$), further aligned with an onboard gyroscope on the MCU. 
During manual motion of the sensor array, the continuous 6-DoF poses captured by the ZED2i--AprilTag system serve as ground truth.
The estimates $(\hat{\bm{R}}, \hat{\bm{p}})$ are compared against ZED2i--AprilTag reference poses $(\bm{R}, \bm{p})$. 
The position error is $e_{\bm{p}} = \|\hat{\bm{p}} - \bm{p}\|_2$, and the orientation error is the geodesic angle on $SO(3)$,
\(e_{\bm{R}} = \arccos((\tr(\hat{\bm{R}}^\top\bm{R})-1)/2)\).
For sublevel dipole localization, the target position error is $e_{\bm{p}_0} = \|\hat{\bm{p}}_0 - \bm{p}_0\|_2$, and the magnetic moment direction error is $e_{\bm{m}_0} = \arccos(\bm{m}_0^\top\hat{\bm{m}}_0/\|\bm{m}_0\|_2\|\hat{\bm{m}}_0\|_2)$, where $\bm{m}_0$ and $\hat{\bm{m}}_0$ represent the ground truth and estimated magnetic moments.

\subsection{Localization Performance}

\subsubsection{6-DoF Array Pose Localization}

\begin{figure}
    \centering
    \includegraphics[width=\columnwidth]{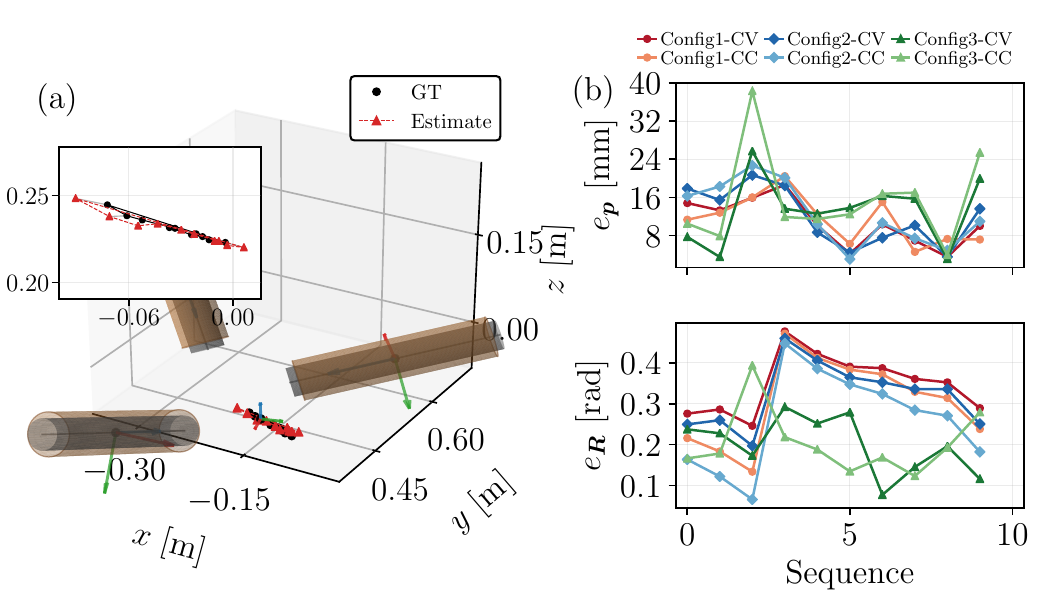}
    \caption{
    Consecutive 6-DoF localization under three sensor array configurations and CV/CC modes. (a) Trajectories. (b) Position and orientation errors.
    }
    \label{fig:6dofresult_cdf}
\end{figure}

\begin{figure}[t]
    \centering
    \includegraphics[width=\columnwidth]{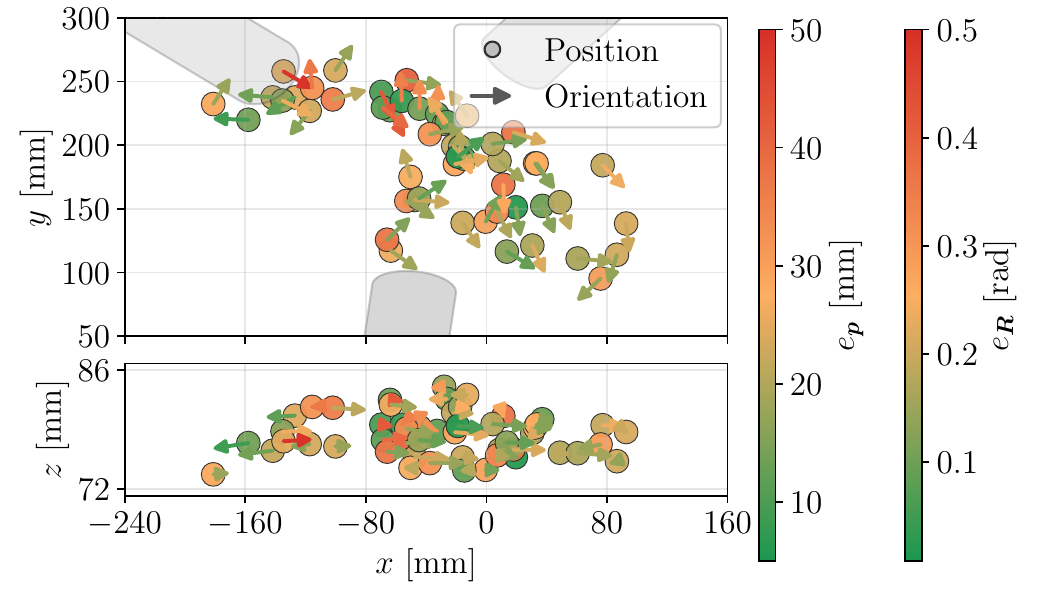}
    \caption{6-DoF localization over random workspace poses using GELS and visual tracking ground truth.}
    \label{fig:6dofresult}
\end{figure}

We first evaluated array pose localization using all 12 sensors (Config.~1), the eight vertex sensors (Config.~2), and a minimum three-sensor subset (Config.~3, No.~2, 3, 12) that satisfies the rank condition $\rank(\mbc D)\geq2$ stated in Sec.~II-C2. 
For each configuration, ten consecutive sequences of slow motion were tested under CV mode using a four slot cycle with background subtraction at a low frequency (\SI{3.47}{\hertz}) and faster CC mode, which omits the background slot and reaches \SI{14.49}{\hertz} in the fastest setting with three sensors. 
As shown in Fig.~\ref{fig:6dofresult_cdf}, the estimated trajectories follow the visual tracking ground truth, with position errors averaged over each sequence of \SI{10.80}{\milli\meter}--\SI{13.22}{\milli\meter} in CV mode and \SI{11.35}{\milli\meter}--\SI{15.57}{\milli\meter} in CC mode.
Across the six settings defined by configuration and mode, standard deviations at the sequence level range from \SI{4.97}{\milli\meter} to \SI{9.87}{\milli\meter}, and the worst position error averaged over a sequence is \SI{38.31}{\milli\meter} for Config.~3 in CC mode.
The best sequence reached \SI{3.07}{\milli\meter}.
These results characterize slow motion under the tested acquisition settings. Noiseless simulations use 12 sensors, fixed orientation, a circle of \SI{40}{\milli\meter} radius, and nominal CV/CC timing. At \SI{10}{\milli\meter\per\second}, mean pose errors across 180 phases are $(1.83~\mathrm{mm},1.00^\circ)$ in CV and $(1.54~\mathrm{mm},1.02^\circ)$ in CC (Supplementary Fig.~3\footnote{Supplementary material: \href{\RALSupplementaryURL}{Supplementary.pdf}. Code and data: \href{\RALRepositoryURL}{\nolinkurl{github.com/WayneWu0526/GELS_code}}.}).

We also tested 60 random poses using Config.~3 under CC mode (see Fig.~\ref{fig:6dofresult}).
The evaluated array centers spanned approximately
$274\times175\times10~\si{\milli\meter}$, while the distances from the
equivalent dipole centers of the three electromagnets to the array center ranged
from \SI{164}{\milli\meter} to \SI{334}{\milli\meter}.
GELS recovered the array pose with average position and orientation errors of \SI{38.93}{\milli\meter} and \SI{0.336}{\radian}, respectively. This larger error reflects the minimum sensor configuration, fast CC operation, and more diverse spatial relationships between sources and the array.
These results show that local field and gradient measurements separated by source yield closed-form 6-DoF pose estimates for both redundant and minimal arrays.

\subsubsection{Sublevel Dipole Localization}

\begin{figure}[t]
    \centering
    \includegraphics[width=\columnwidth]{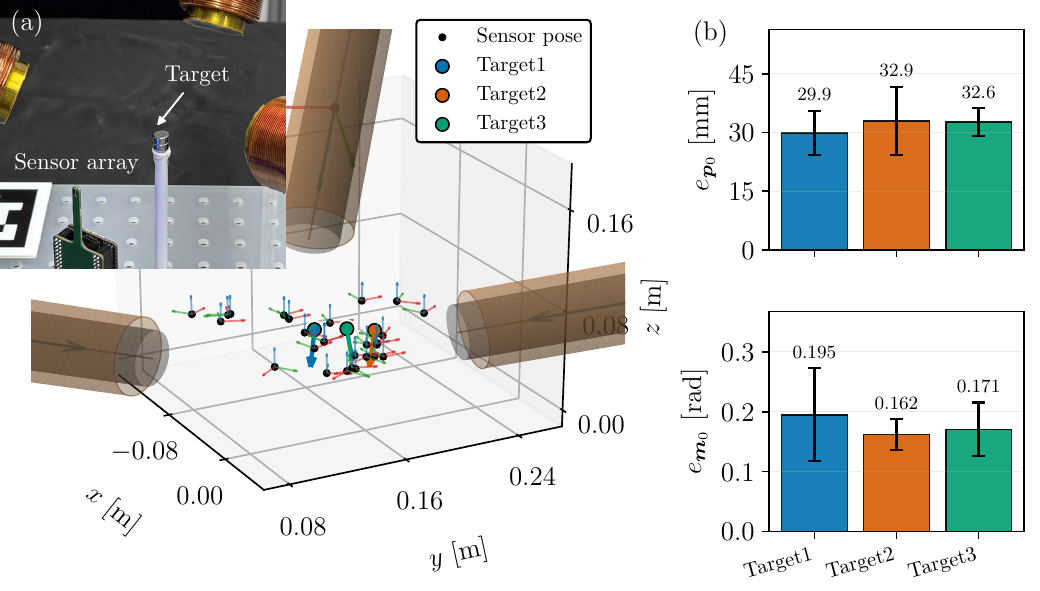}
    \caption{
    Sublevel localization of an external dipole target. (a) Target positions and sensor array poses. (b) Position and magnetic moment direction errors.
    }
    \label{fig:sublevelresult}
\end{figure}

To evaluate sublevel localization using the array pose recovered in CV mode with background subtraction, we tested an external $\Phi\SI{4}{\milli\meter}\times\SI{4}{\milli\meter}$ N52 cylindrical magnet at three locations with ten array poses per target. 
The external dipole position and moment were estimated using Sec.~\ref{subsec:sublevel}, with ground truth obtained from the visual coordinate frame and hole board geometry.
GELS achieved mean position errors of \SI{29.9}{\milli\meter}, \SI{32.9}{\milli\meter}, and \SI{32.6}{\milli\meter}, with magnetic moment direction errors of \SI{0.195}{\radian}, \SI{0.162}{\radian}, and \SI{0.171}{\radian}. 
These results demonstrate that the recovered array pose can serve as a mobile magnetic reference frame for a sublevel dipole localization task.
Because target localization is computed in the recovered array frame, array pose error propagates directly into the target estimate.

Supplementary Table~II compares GELS with representative magnetic localization methods.
In following subsections, we examine how inter-sensor inconsistency, first-order dipole approximation mismatch, and source geometry affect localization accuracy.

\subsection{Influence of Inter-Sensor Measurement Inconsistency}

\begin{figure}[t]
    \centering
    \includegraphics[width=\columnwidth]{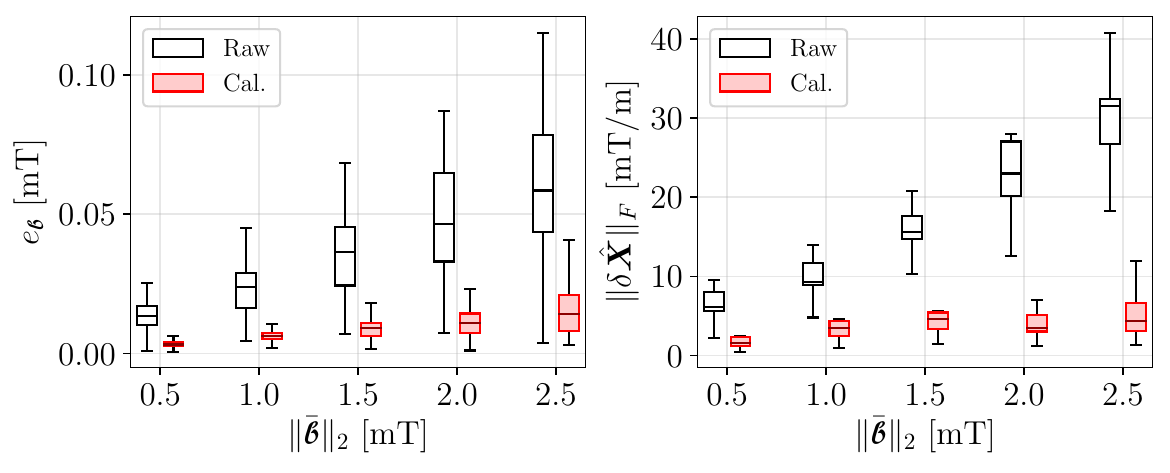}
    \caption{Affine calibration reduces inter-sensor discrepancy and the induced spurious gradient under different uniform magnetic field magnitudes.}
    \label{fig:calib1}
\end{figure}

\begin{figure}[t]
    \centering
    \includegraphics[width=\columnwidth]{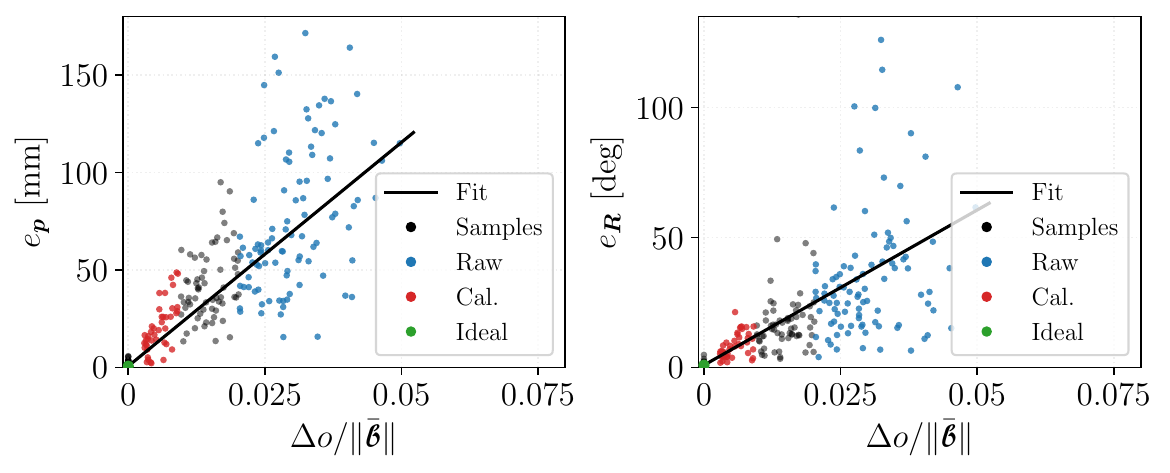}
    \caption{Simulated pose errors versus normalized sensor offset. Raw and Cal. mark illustrative ranges from calibration residuals normalized to $1~\si{\milli\tesla}$.}
    \label{fig:calib2}
\end{figure}

Residual sensor offsets induce a spurious gradient even in a uniform field. For a centered array of fixed shape with a well-conditioned gradient fit, this contribution satisfies $\|\delta\bm X\|_F=\mc O(\Delta o/\ell)$, where $\Delta o$ is the RMS magnitude of centered sensor offsets and $\ell$ is the RMS sensor distance from the array center.

For nonsingular $\bm X$ and small perturbations with $\|\bm X^{-1}\delta\bm X\|_2\ll1$, linearizing $\bm X\bm\rho=-3\bar{\mbc b}$ gives
\begin{equation}
    \label{eqn:perturbation}
    \delta\bm\rho\approx-\bm X^{-1}\lt(\delta\bm X)\bm\rho+3\delta\bar{\mbc{b}}\rt,
\end{equation}
where $\delta\bm\rho$, $\delta\bm X$, and $\delta\bar{\mbc{b}}$ denote displacement, gradient, and field perturbations.
For the gradient contribution $\delta\bm\rho_X$, a well-conditioned dipole gradient then yields
\begin{equation}
    \frac{\|\delta\bm\rho_X\|_2}{\|\bm\rho\|_2}
    =
    \mc O\!\left(
        \frac{\|\bm\rho\|_2}{\ell}
        \frac{\Delta o}{\|\bar{\mbc{b}}\|_2}
    \right),
    \label{eq:rho_error_offset_ratio}
\end{equation}
Thus, for a fixed geometry between sources and the array, displacement error is governed by residual sensor mismatch relative to usable magnetic field magnitude.

Each sensor is calibrated using uniform fields from the Helmholtz coils [Fig.~\ref{fig:expsetup}(b)]. We rotate the array about its fixed local $Z$-axis under separate X-, Y-, and Z-coil excitation at five nominal field magnitudes (\SIrange{0.5}{2.5}{\milli\tesla}). After aligning sensor axes to the array frame, we fit $\mbc b_i^{\mathrm{cal}}=\bm D_i\mbc b_i^{\mathrm{aligned}}+\bm e_i$ by linear least squares. The common reference is the estimated center field, with its magnitude normalized to the mean within each condition. On the calibration data, residual inconsistency decreases from $(38\pm18)~\si{\micro\tesla}$ to $(6\pm3)~\si{\micro\tesla}$, a mean reduction of approximately $85\%$.
Across the uniform field magnitudes tested in Fig.~\ref{fig:calib1}, calibration reduces the mean field discrepancy between each sensor and the array center by $66.9\%$ and the induced spurious gradient by $72.9\%$ on average.

In Fig.~\ref{fig:calib2}, the plotted ratio uses the range of all sensor offset components and the combined source field magnitude $(\sum_k\|\bar{\mbc b}_k\|_2^2)^{1/2}$. The illustrative ideal, Cal., and Raw reference cases have errors $(e_{\bm{p}},e_{\bm{R}})=(5.66~\mathrm{mm},0.53^\circ)$, $(21.46~\mathrm{mm},7.51^\circ)$, and $(107.40~\mathrm{mm},45.51^\circ)$, respectively.

\subsection{Influence of First-Order Dipole Approximation Mismatch}
\label{sec:model_mismatch}

The electromagnet field can deviate from an ideal dipole because of higher-order multipole components determined by coil geometry \cite{petruskaOptimalPermanentMagnetGeometries2013,petruskaOmnimagnetOmnidirectionalElectromagnet2014,abbott2020magnetic}. 
These components introduce a mismatch in the Euler homogeneous relation and propagate to the relative displacement estimate through the perturbation structure in~\eqref{eqn:perturbation}, where $(\delta \bar{\mbc b}, \delta \bm X)$ now denote the higher-order contributions to the local field and gradient, respectively.

\begin{figure}[t]
    \centering
    \includegraphics[width=\columnwidth]{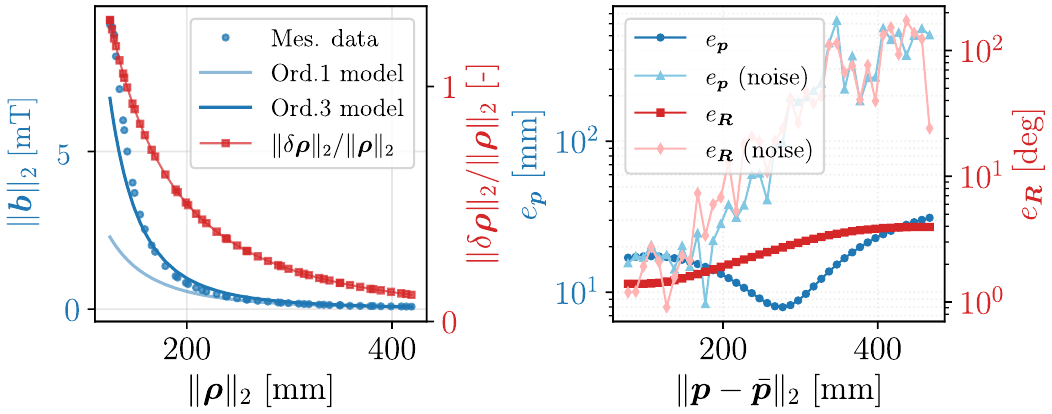}
    \caption{Measured field fits, relative displacement perturbation $\|\delta\bm\rho\|_2 / \|\bm\rho\|_2$, and pose errors versus distance, showing model mismatch and noise effects.}
    \label{fig:highord}
\end{figure}

Fig.~\ref{fig:highord} compares measured fields with first- and third-order fits and reports the induced perturbation $\|\delta\bm\rho\|_2 / \|\bm\rho\|_2$ and pose errors.
At $250\,\mr{mm}$ from the coil center, the order-1 model has a relative field fitting error of $15.40\%$ and a $\delta\bm{\rho}$ ratio of $30\%$.
In the ideal case without noise, $e_{\bm{p}}$ initially decreases as the dipole approximation improves, then slightly increases as the effect of orientation error on position recovery grows with distance from the centroid of the electromagnet sources [see \eqref{eq:position_from_centroids_cf_short}].
With measurement noise, however, $e_{\bm{p}}$ and $e_{\bm{R}}$ are lower at short distances because the field and gradient magnitudes are larger, and grow at larger distances as noise and residual perturbations become larger relative to the useful magnetic signal.
Bounding the ideal dipole gradient and integrating along a sensor baseline of length $\ell$ yields the following necessary condition for a field difference above sensor resolution $\alpha$:
\begin{equation}
    r<\frac{\ell}{2}
    +\left(\frac{3\mu_0\|\bm m_{C_k}\|_2\,\ell}{2\pi\alpha}\right)^{1/4},
    \label{eq:dipole_distance_bound}
\end{equation}
where $r>\ell/2$ is the source distance from the baseline midpoint and $\mu_0$ is the vacuum permeability. This bound gives an upper limit on source distance set by sensor resolution. Using Config.~2 and a \SI{2}{\milli\tesla} reference field, point-dipole simulations assume \SI{0.3}{\micro\tesla} noise standard deviation per axis and \SI{6}{\micro\tesla} RMS residual discrepancy. The finite array approximation also contributes to the total gradient error. Across 200 realizations, increasing the mean source distance from \SI{205.9}{\milli\meter} to \SI{399.0}{\milli\meter} raises the median position error from \SI{10.0}{\milli\meter} to \SI{354.4}{\milli\meter}. Over this range, the gradient signal to total error ratio decreases from \SI{23.3}{\decibel} to \SI{-1.5}{\decibel}. The same trend holds with source moments fitted to measured fields (Supplementary Fig.~2).

\begin{figure}[t]
    \centering
    \includegraphics[width=\columnwidth]{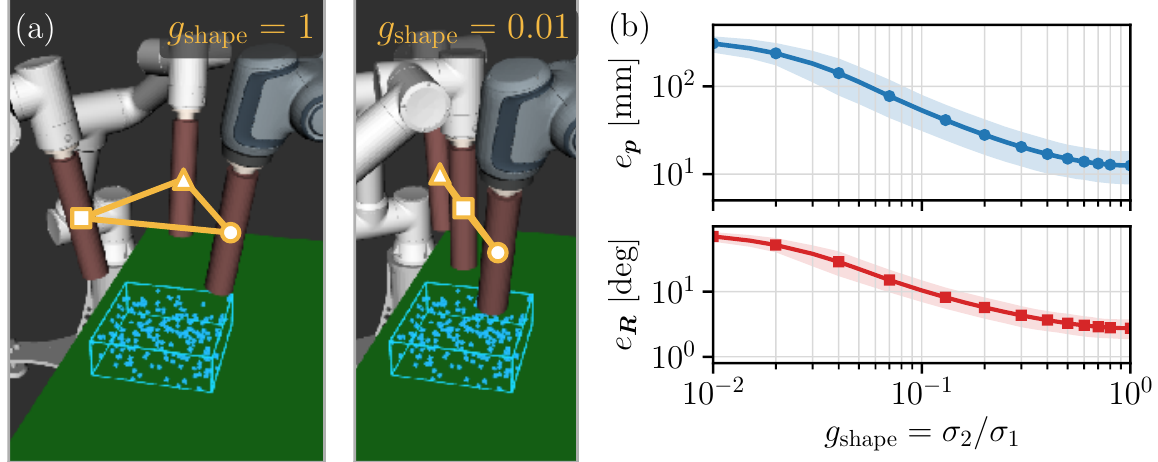}
    \caption{Source geometry and localization.
    (a) Configurations at $g_{\mathrm{shape}}=1$ and 0.01.
    (b) Mean errors and 10th--90th percentile bands across 100 poses in an $80~\si{\milli\meter}$ cube, after averaging 30 noise realizations per pose.}
    \label{fig:source_geometry}
\end{figure}

\Needspace{10\baselineskip}
\subsection{Influence of Electromagnet Source Geometry}
\label{subsec:source_geometry}

For centered source positions
$\bm P_C=[\tilde{\bm p}_{C_1},\ldots,\tilde{\bm p}_{C_M}]$,
we quantify noncollinearity by
\begin{equation}
g_{\mathrm{shape}}
=
\frac{\sigma_2(\bm P_C)}{\sigma_1(\bm P_C)}
\in[0,1],
\label{eq:gshape_metric}
\end{equation}
where $\sigma_1\geq\sigma_2$ are the leading singular values of
$\bm P_C$. For three sources, $g_{\mathrm{shape}}=1$ and $0$ describe
equilateral and collinear layouts, respectively.
Using the experimental array geometry, we simulate virtual dipoles
with independent Gaussian measurement noise of $\SI{0.5}{\micro\tesla}$
standard deviation per axis. We flatten the source layout from
$g_{\mathrm{shape}}=1$ to $0.01$ while fixing its centroid and
$\|\bm P_C\|_F$, preserving the root-mean-square source distance to the centroid.
Fig.~\ref{fig:source_geometry} shows mean pose errors increasing
from $(12.50~\mathrm{mm},2.74^\circ)$ to
$(307.73~\mathrm{mm},70.06^\circ)$ over this sweep. Approaching collinearity
makes the source geometry nearly rank one, weakening the second independent
source direction and amplifying Procrustes orientation errors.

Let $\bm q=\col(\bm q_1,\bm q_2,\bm q_3)$ collect the joint coordinates
and $\bm c(\bm q)=\col(\bm p_{C_1},\bm p_{C_2},\bm p_{C_3})$
the source positions. We propose maximizing $g_{\mathrm{shape}}(\bm P_C(\bm q))$
over $\bm q\in Q$ through the projected ascent update
\begin{equation}
\bm q^{(r+1)}
=
\proj_{Q}
\!\left[
\bm q^{(r)}
+
\eta\bm J_C(\bm q^{(r)})^\top
\nabla_{\bm c}g_{\mathrm{shape}}
\right],
\label{eq:gshape_joint_update}
\end{equation}
where $\eta>0$ is the step size,
$\bm J_C=\partial\bm c/\partial\bm q$, and $Q$ enforces joint limits,
collision avoidance, and minimum source separation. Supplementary Sec.~I details the protocol and simulations of five planned layouts (Supplementary Fig.~1).

\section{Conclusion and Discussion}
\label{sec:conclusion}

This letter presented GELS, a gradiometer-based electromagnetic localization system that converts field and gradient measurements separated by source into closed-form 6-DoF pose estimates.
The central contribution is the sensing construction that obtains local field and gradient pairs for each source from a compact array, allowing Euler displacement recovery and Procrustes registration for an unknown sensor pose under multiple dynamic magnetic sources.
Benchtop experiments report position errors averaged over each sequence of \SI{10.80}{\milli\meter}--\SI{15.57}{\milli\meter}, a median solver runtime of \SI{172.00}{\micro\second}, and a demonstration of sublevel dipole localization.

With known source positions and array geometry, GELS uses no pre-calibrated workspace field maps, initial pose guesses, iterative pose optimization, or calibrated source magnetic moments, while still relying on standard geometric and sensor calibration.
The present accuracy is mainly limited by inter-sensor inconsistency, magnetic signal quality, and dipole model error. While the resulting accuracy is insufficient for precision procedures such as ablation, it may support coarse localization or initialization when mapping is impractical.

The prototype uses TDM to separate measurements from individual sources and may serve as a localization module for manually or mechanically actuated interventional instruments. For magnetically actuated instruments, future implementations could separate localization and actuation field contributions through frequency domain or orthogonal code separation, thereby supporting both functions simultaneously \cite{arxSimultaneousLocalizationActuation2023,vonarxOrthogonalPulseWidthModulationCombined2025}.
Future work will further improve localization performance through integrated gradiometer fabrication, lower inter-sensor inconsistency, an extended usable magnetic signal range, and sources closer to ideal dipoles, with the aim of increasing the range and robustness of closed-form localization.

\bibliographystyle{ieeetr}
\bibliography{Bibliography/references}

\end{document}